\documentclass[10pt,twocolumn,letterpaper]{article}

\usepackage{iccv}   

\definecolor{iccvblue}{rgb}{0.21,0.49,0.74}
\usepackage[pagebackref,breaklinks,colorlinks,allcolors=iccvblue]{hyperref}
\usepackage{booktabs} 
\usepackage{multirow} 
\usepackage{graphicx} 
\usepackage{colortbl} 
\usepackage{xcolor} 
\usepackage[accsupp]{axessibility} 

\def\paperID{12026} 
\def\confName{ICCV}
\def\confYear{2025}

\title{SL$^{2}$A-INR: Single-Layer Learnable Activation for Implicit Neural Representation}

\author{
Moein Heidari\thanks{Equal contribution.}\\
University of British Columbia
\and
Reza Rezaeian\footnotemark[1]\\
University of Tehran
\and
Reza Azad\\
RWTH Aachen University
\and
Dorit Merhof\\
University of Regensburg
\and
Hamid Soltanian-Zadeh\thanks{Corresponding author. Email: hszadeh@ut.ac.ir.}\\
University of Tehran
\and
Ilker Hacihaliloglu\thanks{Corresponding author. Email: ilker.hacihaliloglu@ubc.ca.}\\
University of British Columbia
}

\begin{document}
\maketitle
\begin{abstract}
Implicit Neural Representation (INR), leveraging a neural network to transform coordinate input into corresponding attributes, has recently driven significant advances in several vision-related domains. However, the performance of INR is heavily influenced by the choice of the nonlinear activation function used in its multilayer perceptron (MLP) architecture. To date, multiple nonlinearities have been investigated, but current INRs still face limitations in capturing high-frequency components and diverse signal types. We show that these challenges can be alleviated by introducing a novel approach in INR architecture. Specifically, we propose SL$^{2}$A-INR, a hybrid network that combines a single-layer learnable activation function with an MLP that uses traditional ReLU activations. Our method performs superior across diverse tasks, including image representation, 3D shape reconstruction, and novel view synthesis. Through comprehensive experiments, SL$^{2}$A-INR sets new benchmarks in accuracy, quality, and robustness for INR.
Our code is publicly available at~\href{https://github.com/Iceage7/SL2A-INR}{\textcolor{magenta}{GitHub}}.
\end{abstract}    
\section{Introduction}
\label{sec:intro}
\begin{figure}[!h]
    \centering
    \includegraphics[width=0.5\textwidth , keepaspectratio]{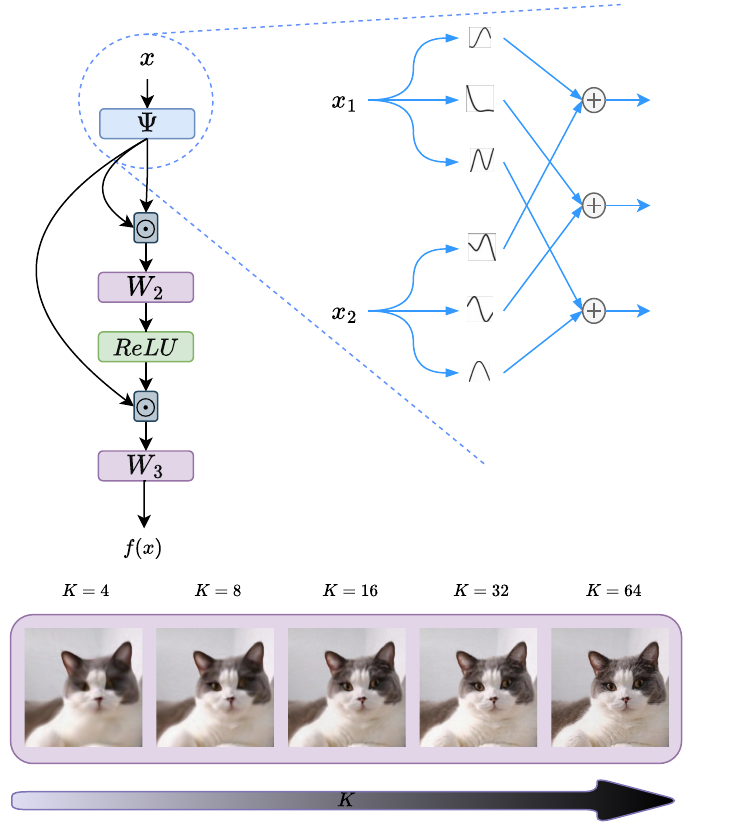}
\caption{An implicit neural representation method that enables flexible tuning of spectral bias through SL$^2$A-INR. The top diagram illustrates the network architecture, which begins with a Learnable Activation Block ($\Psi$) parameterized by Chebyshev polynomials of degree $K$, followed by a feature fusion block. This fusion block is implemented as a vanilla feed-forward network, where the input to each layer is modulated by the output of the first block ($\Psi$). The bottom panel provides a qualitative comparison of reconstruction quality. As the polynomial degree $K$ increases from 4 to 64, the results show a significant enhancement in detail and expressive power, highlighting the improved capacity for finer representations with larger $K$ values. Image adapted from~\cite{chen2022cross}.}

    \label{fig:teaser}
\end{figure}

Implicit neural representations (INRs), which model continuous functions over discrete data points, have recently garnered significant attention for their ability to effectively represent 2D images, 3D shapes, neural radiance fields, and other complex structures~\cite{mildenhall2021nerf, sitzmann2020implicit, park2019deepsdf,shi2024improved}.
Unlike traditional discrete grid-based methods, INRs focus on training neural networks, particularly Multilayer Perceptrons (MLPs), that utilize continuous element-wise nonlinear activation functions to map continuous input coordinates to their respective output values, thereby delivering a continuous signal representation~\cite{sitzmann2020implicit}. Due to their flexible, compact, and efficient design, INRs have garnered immense interest in solving tasks dealing with complex and high-dimensional data. They have demonstrated versatility across a wide range of applications, including computer graphics~\cite{mildenhall2021nerf, muller2022instant, chen2024far}, computer vision~\cite{maiya2023nirvana, zhu2024disorder}, inverse problems~\cite{sun2021coil}, and signal representations~\cite{xu2022signal, sitzmann2020implicit}.
In contrast to Convolutional Neural Networks (CNNs), INRs are not constrained by locality biases, thus exhibit exceptional generalization abilities, feature a concise formulation, and display enhanced flexibility. However, their potential is constrained by various obstacles that limit the application of INRs. Conventional neural networks such as ReLU-based MLPs struggle to capture high-frequency details accurately in signals, such as intricate textures in images or complex shapes in 3D reconstruction, due to their inherent bias towards learning simpler, lower-frequency patterns, known as spectral or low-frequency bias~\cite{rahaman2019spectral, xu2018understanding}. The pioneering efforts in mitigating this issue can be roughly grouped into several main categories. One approach involves increasing the complexity of the input data manifold and explicitly extracting high-frequency features, such as by positional encoding~\cite{tancik2020fourier, muller2022instant}.  An alternative major category of methods utilizes complicated activation functions to achieve more precise approximations~\cite{sitzmann2020implicit,fathony2020multiplicative,saragadam2023wire}. Despite promising results, these approaches still exhibit some degree of spectral bias~\cite{yuce2022structured}, primarily learning faster lower-frequency components of the input. This indicates that there is still room for improvement in addressing spectral bias more comprehensively~\cite{cai2024batch}. Moreover, periodic activation functions are particularly vulnerable to initialization methods, experiencing notable performance declines when not strictly following initialization protocols~\cite{ramasinghe2022beyond}. Therefore, these approaches have difficulty representing fine details in complex shapes, limiting their flexibility across signal types and inverse problems. 
To overcome these challenges, we propose a novel approach that significantly boosts the hierarchical representation capabilities of INRs. Our method achieves superior performance in high-fidelity reconstructions across a wide array of tasks, including images and intricate 3D structures. Additionally, it addresses complex problems such as Neural Radiance Fields (NeRF).
\\
Specifically, the common behavior of MLP networks in the early layers is for neurons to select low-frequency features in the data, which causes higher frequencies to be learned later in the optimization process~\cite{rahaman2019spectral}. Inspired by this, we hypothesize that an accurate polynomial approximation of activation functions in the initial layer of INRs will be sufficient for capturing the high-frequency, fine-grained details and that the spectral spread is influenced by the magnitude of higher-order polynomial coefficients~\cite{mehmeti2020ringing}. To achieve this, as shown in \cref{fig:teaser}, we utilize the stable
and non-oscillatory Chebyshev approximation method and propose a \textbf{\textit{Learnable Activation (LA)}} Block which allows for the capture of finer details and higher frequency components in the function being approximated, while being able to perform spectral-bias adjustment using the Chebyshev coefficients adaptively. Furthermore, we modulate the high-frequency signal passing through the network by adding skip connections to the subsequent ReLU-based MLPs in our \textbf{\textit{Fusion Block}}. This arrangement empowers the network to integrate elements to capture a broad spectrum of frequencies in the input signal, ultimately addressing the challenge of spectral bias.
Our comprehensive experiments across diverse applications clearly showcase the superiority of our approach in robustness, accuracy, quality and convergence rate.
\paragraph{Contributions} 
In summary, this paper offers the following contributions:
\begin{itemize} 
\item We introduce a novel implicit neural representation with learnable activation functions, which allows for capturing finer details and higher frequencies.
\item We conduct extensive experiments across various INR tasks, showing that our approach is effective in capturing high-frequency details compared to other methods.
\item We provide a comprehensive experimental analysis that offers insights into the specifics and design choices of our method.
\end{itemize}
\section{Related Works}
\label{sec:2_related_works}

\textbf{Implicit Neural Representation}
Recent studies have demonstrated significant success in utilizing neural networks to represent various types of signals. These representations are applied across a range of applications, including images \cite{ramasinghe2022beyond,saragadam2023wire,sitzmann2020implicit}, occupancy volumes \cite{ortiz2022isdf}, view synthesis \cite{mildenhall2021nerf,bian2023nope,gao2022nerf}, and audio/video signals \cite{su2022inras,chen2022videoinr}. However, the widely-used ReLU activation function has limitations in learning low-frequency information, which can affect the performance of standard MLPs \cite{tancik2020fourier}. In this context, the ReLU activation function often fails to accurately represent the signal, resulting in the loss of important high-frequency components and leading to a suboptimal overall representation. To address this issue, several solutions have been proposed in the literature. One approach involves encoding the input data with sinusoids \cite{tancik2020fourier}, which maps the input coordinates to a higher dimensional space. The output of this positional encoding is then used as input to the MLP. Another line of research focuses on modifying the activation functions of the MLP. For instance, using periodic activations, as in SIREN \cite{sitzmann2020implicit}, Gaussian activations \cite{ramasinghe2022beyond}, wavelet-based activations \cite{saragadam2023wire}, or variable-periodic activation functions \cite{liu2024finer} can help capture high-frequency details more effectively. However, these methods also come with their own challenges. For example, the representational capacity of SIREN can be highly dependent on the choice of hyperparameters for sinusoidal functions, such as frequency, and can be sensitive to initialization, necessitating careful design choices to avoid random variations \cite{vonderfechtpredicting}.
\\
\textbf{Learnable Activation Functions}
As previously noted, the methods for mitigating spectral bias in INRs have introduced new challenges. In particular, networks utilizing custom activation functions, such as periodic activations, are highly sensitive to initialization schemes and hyperparameter choices. Specifically, the inherent periodicity of sine functions can cause oscillations and slow down convergence, creating challenges during the training process \cite{liu2024finer}. Additionally, despite advancements, some degree of spectral bias remains present in these approaches. Consequently, it is critical to develop an adaptive function to handle nonlinearity and complex frequency distribution effectively.
Instead of relying solely on custom-designed activation functions, which may have their own limitations, we propose using learnable activation functions that adapt based on the data itself as part of our network. A learnable activation function is a type of activation function in neural networks that is not fixed but is optimized during training. Unlike traditional activation functions, which are predetermined and static, learnable activation functions have parameters that are adjusted through backpropagation \cite{goyal2019learning,bingham2022discovering}. These functions can be represented by various forms, such as polynomials, splines, or sigmoid linear units \cite{goyal2019learning,fakhoury2022exsplinet,ramachandran2017searching}, and are designed to adapt dynamically to the data and learning process, potentially enhancing the network's ability to capture complex patterns and improve performance.
Specifically, the recent Kolmogorov-Arnold Network (KAN) \cite{liu2024kan} utilizes B-splines for their activation functions. Capitalizing on this line of research, we transform the input coordinates into a high-dimensional space by projecting them onto multiple adaptive Chebyshev polynomial \cite{ss2024chebyshevpolynomialbasedkolmogorovarnoldnetworks} functions of a fixed degree, departing from the fixed activation functions or positional encodings. These polynomials act as learnable activations, allowing the network to adaptively capture a wide spectrum of frequency information by leveraging the spectral properties of Chebyshev polynomials to efficiently represent both low and high-frequency components of the input signal (as they exhibit progressively higher-frequency oscillatory components in higher degrees). Besides, we incorporate a feature interaction mechanism where the Learnable Activation (LA) Block's output serves as a dynamic modulator for subsequent layers, enhancing the model's ability to capture and propagate high-frequency features throughout the network. For simplicity, we use SL$^{2}$A and SL$^{2}$A-INR interchangeably throughout the paper to refer to our model.

\section{Method}
\label{sec:3_method}


\subsection{Preliminaries}

Implicit Neural Representations (INRs) provide an alternative to traditional discrete data structures, such as pixels or voxels, by encoding information within the weights and structure of a neural network. Formally, an INR can be represented as a neural network \( f_{\theta}: \mathbb{R}^{d_0} \to \mathbb{R}^{d_{L}} \) with \( L \) layers, where \( \theta \in \mathbb{R}^M \) denotes the network parameters. The input \( \mathbf{x} \in \mathbb{R}^{d_0} \) typically corresponds to coordinates (e.g., spatial coordinates in an image), while the output \( \mathbf{y} \in \mathbb{R}^{d_L} \) represents the corresponding signal values (e.g., intensity values in an image).  The network is trained by optimizing a loss function, such as the mean squared error (MSE) loss, which is defined as:

\begin{equation}
\mathcal{L}(\theta) = \frac{1}{N} \sum_{i=1}^{N} \left( f_{\theta}(\mathbf{x}_i) - \mathbf{y}_i \right)^2,
\label{eq:loss}
\end{equation}
where \( N \) denotes the number of training samples. As a continuous representation, INRs offer substantial advantages, including memory efficiency, as they can generate outputs at arbitrary resolutions by querying the continuous function, making them particularly suitable for representing high-dimensional signals, such as 3D scenes. Despite these benefits, traditional neural networks using ReLU activations face a significant challenge known as spectral bias. This refers to the tendency of neural networks to first learn the lower-frequency components of a signal, which limits their ability to effectively capture fine details, particularly high-frequency components. 
\\
Activation functions can be approximated using polynomial expansions, such as the representation of ReLU by Chebyshev polynomials \cite{mehmeti2020ringing}. However, the coefficients of these expansions tend to decay rapidly, which, as shown through harmonic analysis \cite{yuce2022structured}, contributes to spectral bias in neural networks. To address this issue, recent approaches have introduced hand-crafted novel activation functions \cite{saragadam2023wire, liu2024finer}. In contrast, we propose an alternative approach where the polynomial coefficients are learned directly, enabling the network to model highly flexible and complex activation functions. This method effectively mitigates spectral bias while maintaining the expressiveness required for complex tasks.
\subsection{SL$^{2}$A}
To overcome the challenges of spectral bias in neural networks, we introduce a two-block architecture designed to achieve an effective balance between flexibility and efficiency. The first block, termed the Learnable Activation (LA) Block, utilizes learnable activation functions that parameterize high-degree polynomials. This block mitigates the rapid decay of coefficients by allowing higher-degree terms to be learned, thereby enabling flexible spectral bias tuning and enhancing the network’s ability to capture high-frequency components in the data. The second block, termed the Fusion Block, is a ReLU-based network focused on efficient computation (by employing low-rank linear layers). This block complements the LA Block by providing a tradeoff between computational efficiency and expressive power. Additionally, we incorporate a skip connection between the LA Block and the Fusion Block, which enriches the overall representational capacity by directly linking the detailed high-frequency representation from the LA Block to the Fusion Block. This skip connection enhances the model’s ability to represent complex signals effectively, resulting in improved performance and enhanced expressive power.

\subsubsection{Learnable Activation (LA) Block}
Drawing inspiration from the Kolmogorov-Arnold representation theorem \cite{kolmogorov1961representation} and a recent study \cite{liu2024kan} that explores learnable activations applied to edges rather than nodes, we introduce the first block, where activations are parameterized using high-degree polynomials. This enables effective learning of complex activations. The Learnable Activation (LA) block is defined as \( \Psi: \mathbb{R}^{d_0} \to \mathbb{R}^{d_1} \), and can be written in matrix form as:
\begin{equation}
\Psi(\mathbf{x}) = \begin{pmatrix}
\psi_{1,1}(\cdot) & \psi_{1,2}(\cdot) & \cdots & \psi_{1,d_0}(\cdot) \\
\psi_{2,1}(\cdot) & \psi_{2,2}(\cdot) & \cdots & \psi_{2,d_0}(\cdot) \\
\vdots & \vdots & \ddots & \vdots \\
\psi_{d_1,1}(\cdot) & \psi_{d_1,2}(\cdots) & \cdots & \psi_{d_1,d_0}(\cdot)
\end{pmatrix} \mathbf{x},
\end{equation}
where each \( \psi_{i,j}: \mathbb{R} \to \mathbb{R} \) is a learnable activation function, derived from the Chebyshev expansion of the first kind with degree \( K \) \cite{ss2024chebyshevpolynomialbasedkolmogorovarnoldnetworks}. Specifically, the expansion for each \( \psi_{i,j}(x) \) is given by:
\begin{equation}
\psi_{i,j}(x) = \sum_{k=0}^{K} a_{i,j,k} T_k(\sigma(x)),
\end{equation}
where \( T_k: [-1, 1] \to [-1, 1] \) represents the Chebyshev polynomial of the first kind of degree \( k \), and \( a_{i,j,k} \in \mathbb{R} \) are the learnable coefficients, with \( \sigma(x): \mathbb{R} \to (-1, 1) \) chosen to be \( \tanh(x) \). These coefficients \( a_{i,j,k} \) are optimized during backpropagation, enabling the network to adjust the polynomial terms for more effective activation functions during training. As an implementation detail, we apply layer normalization \cite{ba2016layer} immediately after this block to stabilize training \cite{ss2024chebyshevpolynomialbasedkolmogorovarnoldnetworks}. Furthermore, the coefficients \( a_{i,j,k} \) are initialized using the Xavier uniform initialization scheme \cite{glorot2010understanding}, ensuring better convergence during training.

\subsubsection{Fusion Block}
The Fusion Block is defined as a multi-layer perceptron (MLP) composed of several hidden layers, where each layer performs a linear transformation followed by a non-linear activation function, such as ReLU. To enhance the representational capacity of the network, we modulate the input at each layer by the output of the initial Learnable Activation (LA) Block.
\\
Formally, let \( \Psi(\mathbf{x}) \) represent the output of the LA Block given the input \( \mathbf{x} \), and define the layer-wise transformations as follows:
\begin{equation}
\begin{aligned}
\mathbf{z}_1 &= \Psi(\mathbf{x}), \\
\mathbf{z}_l &= \phi\left(\mathbf{W}_l (\mathbf{z}_{l-1} \odot \mathbf{z}_1) + \mathbf{b}_l\right), \quad l = 2, 3, \ldots, L-1, \\
f_{\theta}(\mathbf{x}) &= \mathbf{W}_L (\mathbf{z}_{L-1} \odot \mathbf{z}_1) + \mathbf{b}_L,
\end{aligned}
\end{equation}
where \( \phi \) denotes the ReLU activation function, \( \mathbf{W}_l \in \mathbb{R}^{d_{l-1} \times d_l} \) represents the weights of layer \( l \), and \( \mathbf{b}_l \in \mathbb{R}^{d_l} \) represents the biases of layer \( l \). We adopt lower-rank layers in linear transformations to balance efficiency and performance. The output \( \Psi(\mathbf{x}) \) of the LA Block modulates each layer in the Fusion Block via the element-wise product \( \odot \), ensuring that the influence of the LA Block persists throughout. The final output \( f_{\theta}(\mathbf{x}) \) of the Fusion Block is then computed by applying a linear transformation and bias to the modulated output of the last hidden layer \( \mathbf{z}_{L-1} \).
\\
This modulation improves the performance of the model by dynamically adjusting the influence of each hidden layer based on the initial Learnable Activation Block. This dynamic adjustment helps in capturing more intricate patterns and reduces spectral bias, ultimately leading to enhanced representation capabilities and better overall performance. 
\\
A simpler architecture involves an MLP without modulation by the LA Block output, where information flows in a standard feed-forward manner. However, as our ablation study shows, this non-modulated approach significantly reduces performance, highlighting the importance of our proposed design.

\begin{figure*}[!ht]
    \centering
    \includegraphics[width=0.9\textwidth,keepaspectratio]{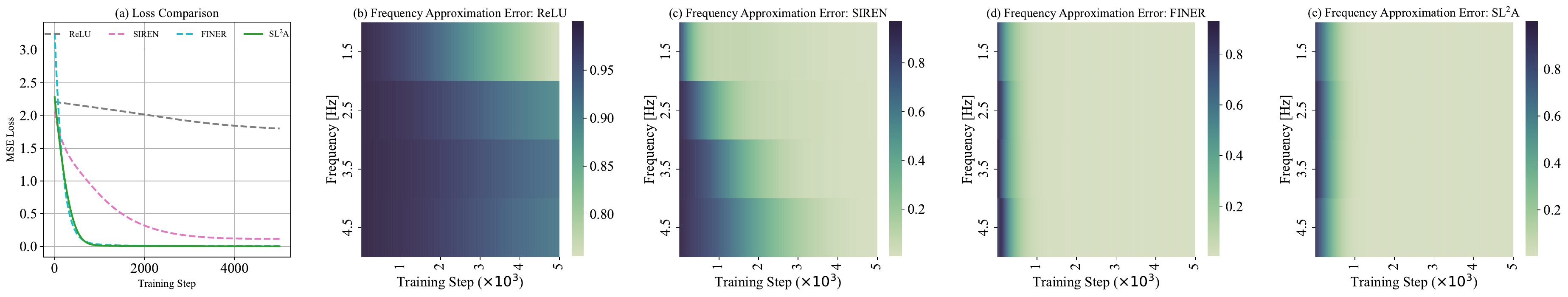}
    \caption{Comparison of convergence rate and frequency approximation error of different methods in 1d function fitting.}

    \label{fig:spectral_bias_cvpr}
\end{figure*}

\begin{table*}
    \scriptsize
    \centering
    \caption{Comparison of different methods based on PSNR (in dB) / SSIM (0-1) for 2D color image approximation results for images 00 to 15 and the average. The table is split into two parts for clarity. Higher values indicate better quality, with the best result in \textcolor{red}{red} and the second best in \textcolor{blue}{blue}.}
    \label{Table:image_div2k_full}
    
    \begin{tabular}{l|c|cccccccc}
        \toprule
        Method & \#Params (K)$\downarrow$ & 00 & 01 & 02 & 03 & 04 & 05 & 06 & 07 \\
        \midrule
        FINER & \textcolor{blue}{198.9} & \textcolor{blue}{32.00} / \textcolor{blue}{0.862} & 36.52 / 0.928 & \textcolor{blue}{35.16} / 0.900 & \textcolor{blue}{38.08} / \textcolor{blue}{0.948} & \textcolor{red}{36.08} / \textcolor{blue}{0.942} & \textcolor{blue}{32.92} / \textcolor{blue}{0.889} & \textcolor{blue}{35.69} / 0.941 & \textcolor{blue}{35.69} / 0.903 \\
        Gauss & \textcolor{blue}{198.9} & 30.08 / 0.847 & 34.72 / 0.920 & 33.72 / 0.890 & 35.21 / 0.908 & 34.53 / 0.941 & 31.33 / 0.862 & 35.37 / \textcolor{blue}{0.949} & 33.65 / 0.870 \\
        ReLU+P.E. & 204.0 & 30.59 / 0.851 & \textcolor{blue}{37.95} / \textcolor{red}{0.954} & 34.61 / \textcolor{red}{0.922} & 37.44 / 0.943 & 33.57 / 0.931 & 31.22 / 0.854 & 33.30 / 0.897 & 34.98 / \textcolor{blue}{0.914} \\
        SIREN & \textcolor{blue}{198.9} & 29.29 / 0.831 & 35.56 / 0.922 & 31.67 / 0.887 & 36.72 / 0.938 & 31.90 / 0.918 & 30.73 / 0.836 & 33.15 / 0.916 & 32.09 / 0.878 \\
        WIRE & \textcolor{red}{91.6} & 28.00 / 0.773 & 31.43 / 0.848 & 29.52 / 0.784 & 31.91 / 0.833 & 29.84 / 0.833 & 29.26 / 0.821 & 31.46 / 0.874 & 29.35 / 0.744 \\
        \midrule
        SL$^{2}$A & 330.2 & \textcolor{red}{33.40} / \textcolor{red}{0.892} & \textcolor{red}{38.99} / \textcolor{blue}{0.950} & \textcolor{red}{35.39} / \textcolor{blue}{0.914} & \textcolor{red}{38.96} / \textcolor{red}{0.951} & \textcolor{blue}{36.06} / \textcolor{red}{0.951} & \textcolor{red}{34.02} / \textcolor{red}{0.903} & \textcolor{red}{36.95} / \textcolor{red}{0.954} & \textcolor{red}{36.45} / \textcolor{red}{0.915} \\
        \bottomrule
    \end{tabular}

    \vspace{1em} 

    \begin{tabular}{l|cccccccc|c}
        \toprule
        Method & 08 & 09 & 10 & 11 & 12 & 13 & 14 & 15 & \textbf{Average} \\
        \midrule
        FINER & \textcolor{red}{32.21} / \textcolor{red}{0.933} & \textcolor{blue}{34.48} / \textcolor{blue}{0.910} & 40.08 / 0.965 & 45.42 / 0.983 & \textcolor{red}{37.83} / \textcolor{red}{0.863} & \textcolor{blue}{35.71} / 0.922 & \textcolor{blue}{37.43} / \textcolor{blue}{0.963} & \textcolor{blue}{36.29} / \textcolor{blue}{0.932} & \textcolor{blue}{36.35} / \textcolor{blue}{0.924} \\
        Gauss & 30.82 / 0.920 & 33.04 / 0.904 & 39.74 / 0.961 & \textcolor{blue}{45.77} / \textcolor{blue}{0.988} & 35.02 / 0.841 & 35.02 / \textcolor{blue}{0.926} & 35.71 / 0.952 & 35.59 / 0.938 & 34.96 / 0.914 \\
        ReLU+P.E. & 29.17 / 0.879 & 32.91 / 0.908 & \textcolor{blue}{40.27} / \textcolor{blue}{0.973} & \textcolor{red}{47.18} / \textcolor{red}{0.989} & \textcolor{blue}{37.62} / \textcolor{blue}{0.856} & 33.30 / 0.897 & 35.64 / 0.937 & 34.59 / 0.947 & 35.27 / 0.916 \\
        SIREN & 29.47 / 0.898 & 30.36 / 0.867 & 37.25 / 0.950 & 42.50 / 0.968 & 35.72 / 0.820 & 32.01 / 0.872 & 34.88 / 0.918 & 32.23 / 0.915 & 33.47 / 0.896 \\
        WIRE & 28.18 / 0.860 & 28.52 / 0.773 & 33.77 / 0.862 & 36.00 / 0.882 & 29.84 / 0.694 & 30.72 / 0.834 & 31.78 / 0.876 & 30.49 / 0.805 & 30.63 / 0.818 \\
        \midrule
        SL$^{2}$A & \textcolor{blue}{31.81} / \textcolor{blue}{0.931} & \textcolor{red}{34.56} / \textcolor{red}{0.914} & \textcolor{red}{41.04} / \textcolor{red}{0.974} & 45.47 / 0.984 & 35.95 / 0.838 & \textcolor{red}{35.74} / \textcolor{red}{0.93} & \textcolor{red}{38.51} / \textcolor{red}{0.970} & \textcolor{red}{36.70} / \textcolor{red}{0.951} & \textcolor{red}{36.88} / \textcolor{red}{0.933} \\
        \bottomrule
    \end{tabular}
\end{table*}
\section{1D Simple Function Approximation}
\label{sec:1d-analysis}
In \cite{rahaman2019spectral}, the authors demonstrate that MLP-based networks exhibit frequency-dependent learning speeds and provide empirical evidence of spectral bias, where lower frequencies are learned first. To investigate the spectral bias in our network, we adapted the experimental setup from \cite{shi2024improved}, using a 1D rounded periodic function with four dominant frequencies in the spectrum, as shown in the following equation:
\begin{equation}
    g(x) = 2R\left( \frac{\sin(3\pi x) + \sin(5\pi x) + \sin(7\pi x) + \sin(9\pi x)}{2} \right),
    \label{eq:rounding_func}
\end{equation}
where \( g(x) \) is defined over 2048 discrete points, uniformly sampled from the interval \([-1, 1]\). In this expression, \( R \) represents the rounding function, which introduces discontinuities into the data and increases the difficulty of training. This setup enables us to analyze how the network handles different frequency components, particularly in the presence of spectral bias.
\\
We conducted experiments using our proposed SL$^{2}$A model with polynomial degrees \( 512 \) and \( L = 3 \) layers. For comparison, we evaluated several additional models, including SIREN \cite{sitzmann2020implicit}, which uses sinusoidal activation functions, and the recently proposed FINER \cite{liu2024finer} activation function. To ensure a fair comparison, each of the alternative models was implemented with \( L = 5 \) layers. In this experiment, we followed the experimental setups from the respective papers for SIREN and FINER, specifically setting \( \omega_0 = 5 \) for SIREN to support fast convergence \cite{shi2024improved, sitzmann2020implicit}, and \( \omega_0 = 1 \) for FINER \cite{liu2024finer}. The initialization schemes from these papers were also followed to ensure optimal performance. 
\\
 \Cref{fig:spectral_bias_cvpr} compares the mean squared error (MSE) loss and frequency approximation error across different models, including ReLU, SIREN, FINER, and our proposed SL$^{2}$A. In \cref{fig:spectral_bias_cvpr}(a), the loss convergence rates demonstrate that SL$^{2}$A achieves faster convergence than SIREN and ReLU, closely matching FINER’s performance. As shown in \cref{fig:spectral_bias_cvpr}(b)-(e), the frequency approximation error over training steps for each model is presented, with a particular focus on analyzing the main frequency components of the input function. ReLU (\cref{fig:spectral_bias_cvpr}(b)) exhibits high spectral bias, with slow learning for higher frequencies, indicating that it primarily focuses on lower frequencies early in training. SIREN (\cref{fig:spectral_bias_cvpr}(c)) mitigates some spectral bias but still struggles with high-frequency approximation, leading to slower convergence for complex signals. FINER (\cref{fig:spectral_bias_cvpr}(d)) and SL$^{2}$A (\cref{fig:spectral_bias_cvpr}(e)) demonstrate significantly lower frequency approximation errors across the entire frequency range, with SL$^{2}$A maintaining a consistently low error for both low and high frequencies. This indicates that SL$^{2}$A can effectively approximate multiple frequency components early in training, reducing the frequency-dependent learning limitations observed in other models. The results validate SL$^{2}$A’s ability to handle high-frequency details efficiently, combining rapid convergence with minimal spectral bias.

\section{Experimental Results}
\label{sec:5_experiments}

To evaluate the effectiveness of our proposed method, we conducted several experiments in signal representation tasks such as 2D image representation, 3D shape reconstruction, and novel view synthesis. Our experiments were conducted using PyTorch on an Nvidia RTX 4070 GPU equipped with 12GB of memory. 

\begin{figure*}[!ht]
    \centering
    \includegraphics[width=0.85\textwidth,keepaspectratio]{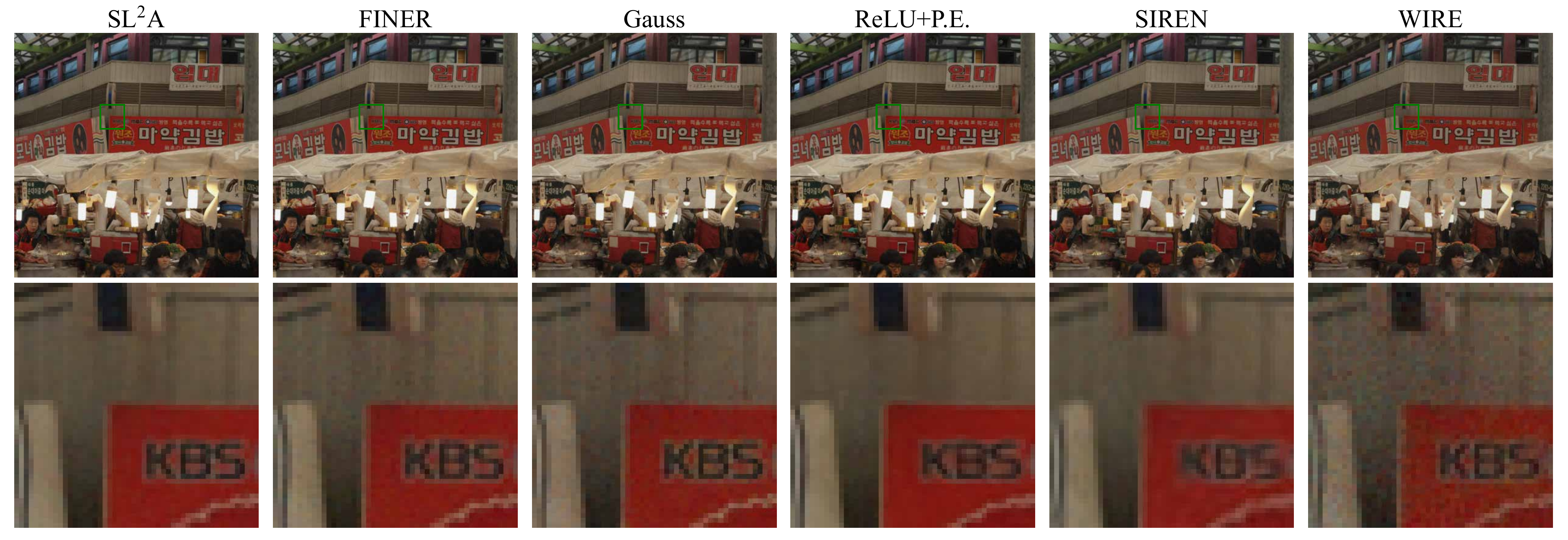}
    \caption{Comparison of image representation of SL$^{2}$A with other methods.}

    \label{fig:image-representation}
\end{figure*}

\begin{figure*}[!ht]
    \centering
    \includegraphics[width=0.8\textwidth, height=10cm, keepaspectratio]{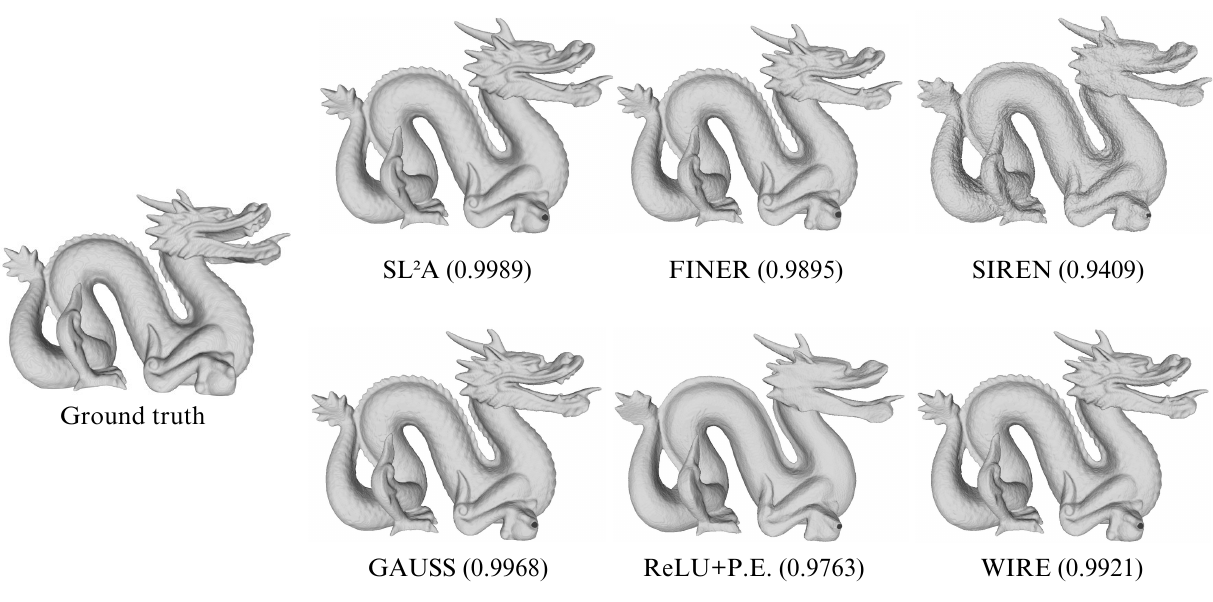}
    \caption{Results for occupancy (dragon shape) with different methods. The numbers in parentheses represent the IoU metric.}
    \label{fig:image-occupancy}
\end{figure*}

\begin{figure*}[!ht]
    \centering
    \includegraphics[width=0.9\textwidth, keepaspectratio]{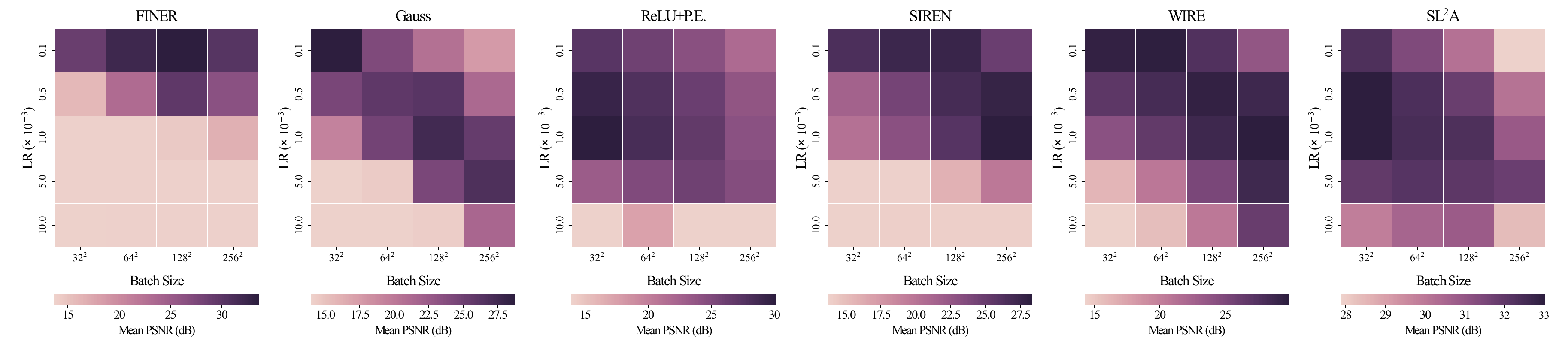}
    \caption{Heatmap of mean PSNR across different batch sizes and learning rates (LR) for various methods on 5 random images from the DIV2K \cite{timofte2017ntire} dataset. Our method, SL\(^2\)A, demonstrates greater robustness to variations in the learning rate and batch size compared to most other methods.}
    \label{fig:heatmap}
\end{figure*}

\subsection{Signal Representations}
\subsubsection{Image Representation}
We first applied our method to the task of 2D image fitting using INR. We explored all possible combinations of learning rates (1.5$\times$10$^{-4}$, 10$^{-4}$, 1.5$\times$10$^{-3}$, 10$^{-3}$, 10$^{-2}$) and batch sizes (32$\times$32, 64$\times$64, 128$\times$128, 256$\times$256). For each combination, we ran each model on five different images \cite{timofte2017ntire}, averaging the results to select the best configuration of learning rate and batch size. The loss function used is the MSE between the model's output and the actual pixel values in the images.

\noindent
\textbf{Data}
For our image representation experiments, we used 16 natural images from the DIV2K dataset \cite{timofte2017ntire}, each with a resolution of $512 \times 512$ \cite{liu2024finer}.

\noindent
\textbf{Results}
We evaluated our method by comparing its performance against five baseline models in terms of Peak Signal-to-Noise Ratio (PSNR) and Structural Similarity Index (SSIM) \cite{hore2010image}. The baselines included WIRE \cite{saragadam2023wire}, SIREN \cite{sitzmann2020implicit}, FINER \cite{liu2024finer}, ReLU+P.E. \cite{tancik2020fourier}, and Gauss \cite{ramasinghe2022beyond}. For consistency, we trained each model for 500 epochs and reported the best results achieved. All models were implemented with a standard architecture of \( L=5 \) layers, each containing 256 neurons, except for WIRE, which, following its original design, used a multilayer perceptron (MLP) with 4 layers and 300 neurons per layer. In our approach, for a comparable number of parameters, we used a Learnable Activation (LA) Block with polynomial degrees set to \( K=256 \), and incorporated linear layers with rank 128 in the Fusion Block followed by ReLU activations. This architecture was designed to balance flexibility and computational efficiency, enabling the model to effectively capture high-frequency details while maintaining efficient computation. The quantitative results for images representation task are shown in \cref{Table:image_div2k_full}. Our model performed best, or at least second-best, in most images, achieving state-of-the-art results. In the few instances where the results were comparable, our approach still demonstrated its effectiveness, achieving competitive performance. Furthermore, our model produced the best average results across all images, highlighting its overall effectiveness.
We also provide qualitative results in \cref{fig:image-representation}, which shows a representative sample image tested across all baseline methods and our model. The figure contains magnified regions of the images where text is present. The zoomed-in portions show that our approach outperforms the baselines by reducing the blurring effect, allowing the text to be represented more clearly.



\subsubsection{Occupancy Volume Representation}
For this task, we maintain the same architectural settings as in the image representation task, except that we set the ranks of the linear layers to 32 for efficiency. We follow the baseline code of \cite{shi2024improved} with the experimental setup used in FINER \cite{liu2024finer}, while our network translates 3D coordinates (\textbf{\textit{M} = 3}) into signed distance function (SDF) values (\textbf{\textit{N} = 1}).
In this experiment, five shapes from the publicly available Stanford 3D Scanning Repository dataset \cite{muller2022instant} are used for evaluation.
\Cref{tab:res_sdf} presents a quantitative comparison between the proposed method and four baseline approaches. Across the evaluated shapes, SL$^{2}$A consistently outperforms the baselines, achieving the best results in all cases. \Cref{fig:image-occupancy} provides comparisons between the proposed method and six baselines showing superior reconstruction quality. Specifically, both low-frequency smooth regions (legs and lower-body curves), and the high-frequency rough parts (dragons face) are well represented by SL$^{2}$A, providing consistent performance as opposed to other methods. 
This balanced representation is attributed to our method's synergistic design, which combines the fusion blocks for capturing low-frequency components with a learnable activation block specialized in high-frequency learning, allowing for a comprehensive spectral representation with the highest IOU values.
\begin{table}
\setlength\tabcolsep{2.5pt}
\scriptsize
\centering
\caption{Quantitative comparisons on representing signed distance field.}
\begin{tabular}{clcccccc}
\toprule
 &{Methods}
& {Armadillo} & {Dragon} & {Lucy} & {Thai Statue} 
& {BeardedMan} \\
\midrule
\multirow{6}*{\rotatebox{90}{IOU $\uparrow$}} 
& {FINER}   & \textcolor{blue}{9.899e-1} & 9.895e-1 & \textcolor{blue}{9.832e-1} & 9.848e-1 & 9.943e-1 \\        
& {Gauss}   & 9.768e-1 & \textcolor{blue}{9.968e-1} & 9.601e-1 & 9.900e-1 & 9.932e-1 \\            
& {ReLU+P.E.} & 9.870e-1 & 9.763e-1 & 9.760e-1 & 9.406e-1 & 9.939e-1 \\          
& {SIREN}   & 9.895e-1 & 9.409e-1 & 9.721e-1 & 9.799e-1 & \textcolor{blue}{9.948e-1} & \\
& {WIRE}    & 9.893e-1 & 9.921e-1 & 9.707e-1 & \textcolor{blue}{9.900e-1} & 9.911e-1 & \\  
& {SL$^{2}$A} & \textcolor{red}{9.983e-1} & \textcolor{red}{9.989e-1} & \textcolor{red}{9.988e-1} & \textcolor{red}{9.986e-1} & \textcolor{red}{9.987e-1} \\   
\bottomrule
\label{tab:res_sdf}
\vspace{-0.5cm}
\end{tabular}
\end{table}

\subsection{Neural Radiance Fields}
\label{sec:nerf}

Neural Radiance Fields (NeRFs) \cite{mildenhall2021nerf} provide a method for synthesizing novel views of complex 3D scenes by representing them as continuous volumetric functions. These functions are optimized using 2D images and corresponding camera poses. The NeRF architecture employs a Multi-Layer Perceptron (MLP) to implicitly encode the scene. It takes 5D input, comprising a 3D spatial location and a 2D viewing direction, and outputs the associated RGB color and volume density. 
For training, we adopt the same configurations as in previous work \cite{liu2024finer} for all baselines and report their corresponding results accordingly. To demonstrate SL$^2$A's capability in capturing high-frequency details, we employ a reduced training set of 25 images rather than the standard 100 images, consistent with previous studies \cite{liu2024finer, saragadam2023wire}.
\Cref{tab:res_nerf} presents a quantitative comparison of our method against various methods on the Blender dataset \cite{mildenhall2021nerf}. SL$^2$A outperforms competing approaches in nearly all scenes and remains highly competitive in the few cases where it does not achieve the best results. This demonstrates SL$^2$A's ability to deliver consistently strong performance across diverse tasks.

\begin{table}
\setlength\tabcolsep{2.5pt}
\scriptsize
\centering
\caption{Quantitative comparisons on novel view synthesis.}
\begin{tabular}{clcccccccc}
\toprule
 &{Methods}
& {Chair} & {Drums} & {Ficus} & {Hotdog} 
& {Lego} & {Materials} & {Mic} & {Ship} \\
\midrule
\multirow{6}*{\rotatebox{90}{PSNR $\uparrow$}} 
& {ReLU+P.E.} & 31.32 & 20.18 & 24.49 & 30.59
                & 25.90 & 25.16 & 26.38 & 21.46 \\
&{Gauss}  & 32.68 & 23.16 & 26.10 & 32.17
                & 28.29 & 26.19 & 33.59 & 22.28 \\
&{SIREN}  & 33.31 & \textcolor{blue}{24.89} & 27.26 & 32.85
                & 29.60 & \textcolor{blue}{27.13} & 33.28 & 22.25 \\
&WIRE       & 29.31 & 22.22 & 25.91 & 30.11 & 25.76 & 25.05 & 32.35 & 21.15 \\                

&{FINER}   & \textcolor{blue}{33.90 }& \textcolor{red}{24.90} & \textcolor{red}{28.70} & \textcolor{blue}{33.05}
                & \textcolor{blue}{30.04} & 27.05 & \textcolor{red}{33.96} & \textcolor{blue}{22.47} \\
&{SL$^{2}$A}   & \textcolor{red}{34.70}    & 24.33    & \textcolor{blue}{28.31 }   & \textcolor{red}{33.83}   & \textcolor{red}{30.63}    & \textcolor{red}{28.62}    & \textcolor{blue}{33.88}    & \textcolor{red}{23.43} \\ 
\bottomrule
\label{tab:res_nerf}
\vspace{-0.5cm}
\end{tabular}
\end{table}

\section{Discussion}
\label{sec:6_analysis}

In this section, we delve into the impact of various design choices on the model’s overall performance. To this end, we conducted a thorough ablation study, examining the effect of polynomials degree, the distribution of eigen values of neural tangent kernel (NTK)  \cite{jacot2018neural} and the role of the Fusion block.
Further details and analysis are provided in the supplementary material.
\subsection{Chebyshev Polynomial Degree and the Role of the Fusion Block}

\begin{figure}[ht]
    \centering
    \begin{subfigure}[b]{0.48\linewidth}
        \centering
        \includegraphics[width=\linewidth]{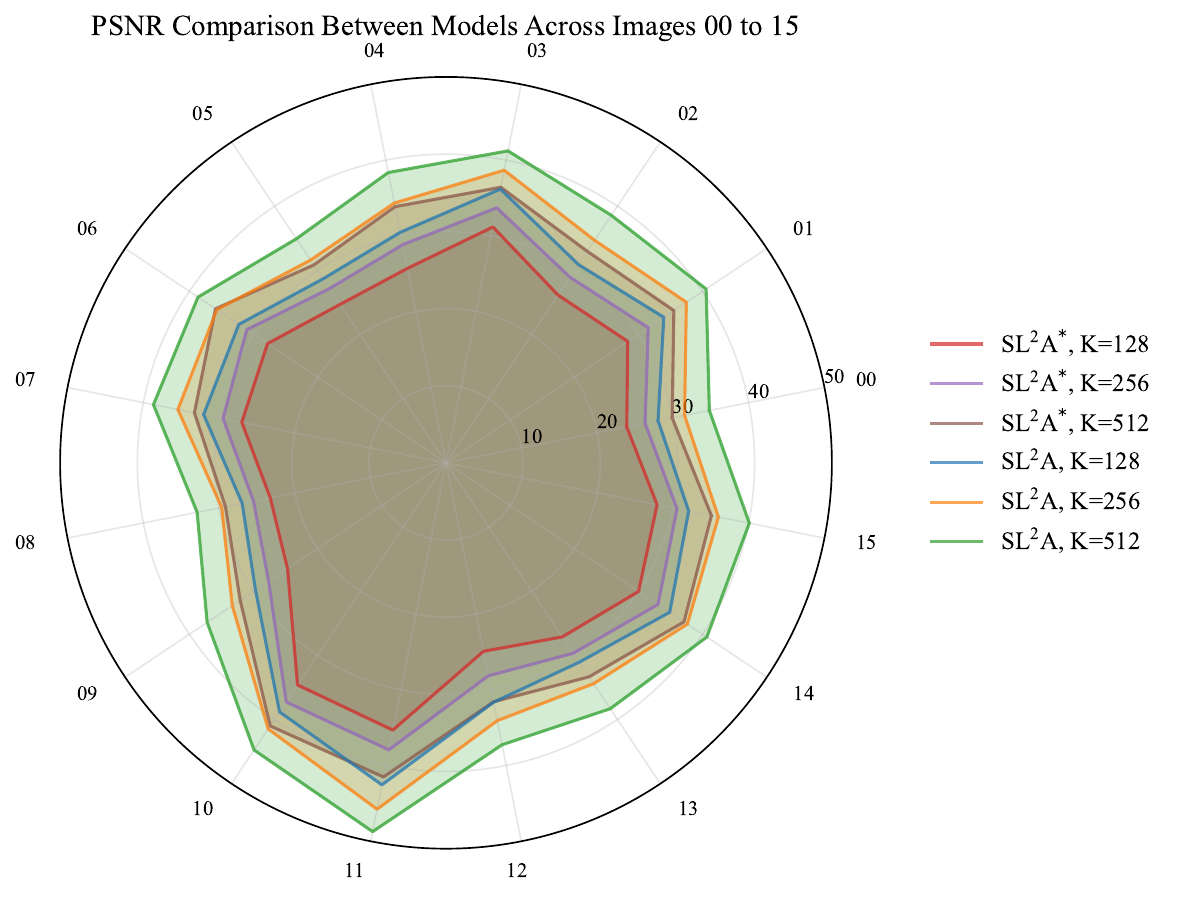}
        \caption{Effect of Chebyshev polynomial degree and skip-connections on PSNR.}
        \label{fig:effect-degree}
    \end{subfigure}
    \hfill
    \begin{subfigure}[b]{0.48\linewidth}
        \centering
        \includegraphics[width=\linewidth]{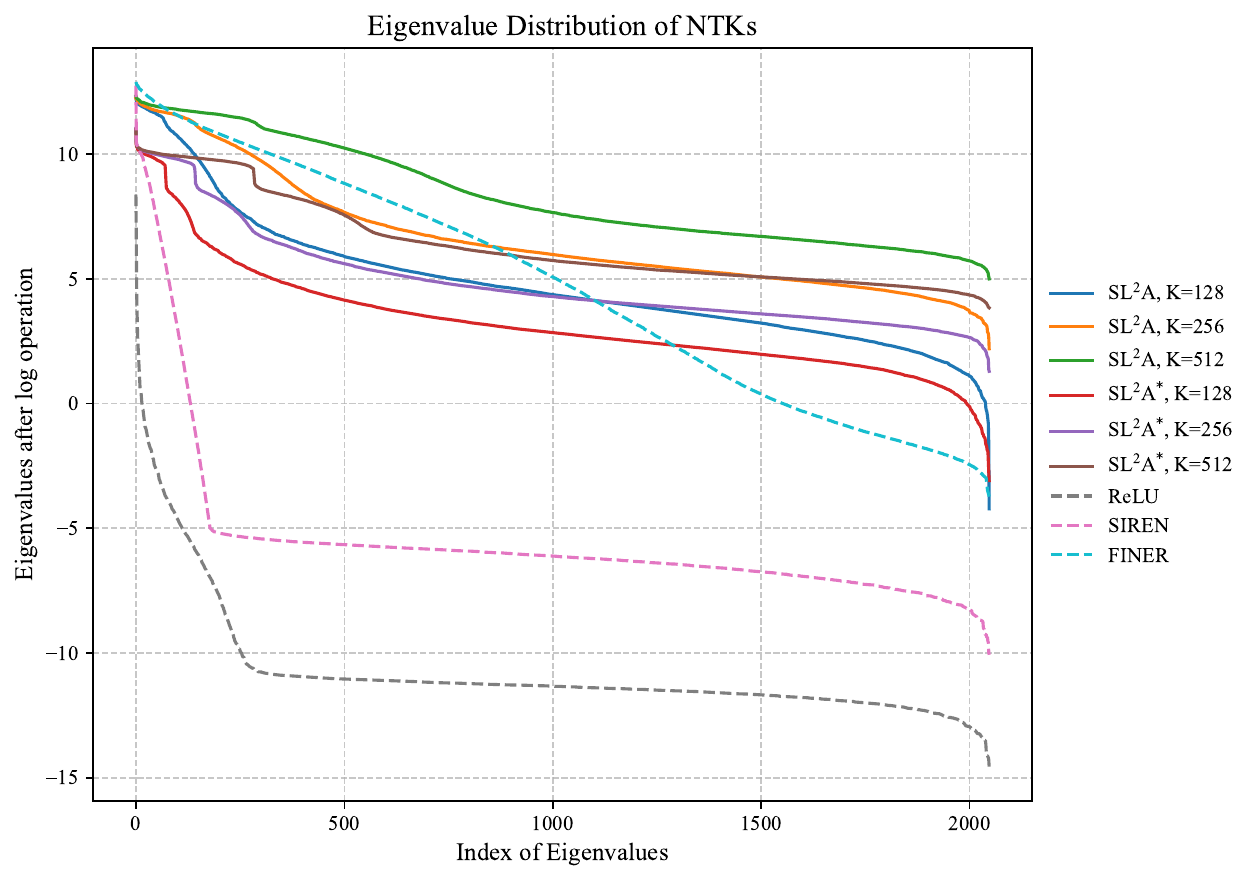}
        \caption{Distribution of eigenvalues of NTK.}
        \label{fig:ntk}
    \end{subfigure}
    \caption{Visualizations comparing ablation results and kernel eigenvalues.}
    \label{fig:combined}
\end{figure}

\cref{fig:effect-degree} demonstrates the impact of different Chebyshev polynomial degrees ($K$) with and without skip-connections (indicated by an asterisk) in our method. The addition of skip-connections shows a significant improvement, enabling the high-frequency information learned in the initial trainable activation layer to propagate more effectively through subsequent layers. Furthermore, increasing the polynomial degree $K$ generally improves performance across images.


\subsection{Neural Tangent Kernel Perspective}
Neural Tangent Kernel (NTK) \cite{jacot2018neural} provides valuable insights into the training dynamics of neural networks, where the convergence properties are influenced by the distribution of NTK eigenvalues. For a neural network \( f(x; \theta) \) with parameters \( \theta \) and input \( x \), the NTK, denoted by \( \Theta(x, x') \), is mathematically defined as the inner product of the gradients of \( f \) with respect to \( \theta \) at two input points \( x \) and \( x' \):
\begin{equation}
\Theta(x, x') = \mathbb{E}_{\theta} \left[ \nabla_{\theta} f(x; \theta) \cdot \nabla_{\theta} f(x'; \theta) \right],
\end{equation}
where \( \nabla_{\theta} f(x; \theta) \) represents the gradient of the network output with respect to its parameters, and the expectation is taken over the random initialization of \( \theta \). The NTK eigenvalues play a crucial role in determining the rate at which different components of the target function are learned. Components corresponding to larger kernel eigenvalues are learned faster, which is important for overcoming spectral bias in neural networks \cite{yuce2022structured, tancik2020fourier}.
\noindent
In \cref{fig:ntk}, we present the NTK eigenvalue distribution \cite{yuce2022structured} for various models, focusing on the 1D input-output case described in \cref{sec:1d-analysis}. Increasing \( K \) in our method reduces the rate of eigenvalue decay, resulting in higher values that enhance the model's ability to capture high-frequency components, which is beneficial for faster learning. Removing skip connections (denoted by asterisk) accelerates eigenvalue decay, highlighting their role in preserving spectral properties. As shown in \cref{fig:ntk}, we observe a hierarchy in decay patterns: ReLU exhibits the most rapid decay, followed by SIREN, which maintains stronger spectral properties, and FINER, which decays more gradually but still faster than our method. 
\noindent
\subsection{Robustness to Hyperparameter Settings}
\Cref{fig:heatmap} presents a heatmap of PSNR values across a range of batch sizes and learning rates (LR) for various INRs. As shown, INRs generally exhibit sensitivity to these hyperparameters, with performance varying significantly across different settings. However, our SL\(^2\)A displays greater stability and achieves higher PSNR values across a broader range of hyperparameter configurations, especially when compared to alternative methods like FINER, Gauss, and SIREN. This robustness suggests that SL\(^2\)A is less affected by the choice of hyperparameters, making it more adaptable to different training conditions.
\section{Limitations and Future Work}
SL$^{2}$A builds on prior work (KAN) \cite{liu2024kan}, improving its efficiency by using Chebyshev polynomials in the first layer and incorporating a hybrid architecture with low-rank linear layers. However, scaling it to very large architectures, as inherited from KANs, can be costly. Moreover, in the image-fitting task, SL$^{2}$A introduces additional, though small, parameters. Incorporating sparsification or weight sharing represents a promising future direction to address these challenges.

\section{Conclusion}
We introduced and validated SL$^2$A, a novel INR approach that enhances frequency representation. Unlike methods relying on handcrafted activation functions, SL$^2$A incorporates a learnable activation layer enabling flexible spectral bias tuning and effective feature modulation in subsequent ReLU-based MLP layers, thus capturing high-frequency details more effectively. This design minimizes extensive hyperparameter tuning and demonstrates robust performance across various scenarios. Extensive evaluations confirm that SL$^2$A consistently outperforms existing INR methods on image fitting, 3D shape representation, and neural rendering tasks, as detailed comprehensively in the main paper and supplemental materials.

\section*{ACKNOWLEDGMENTS}
This work was supported by the Canadian Foundation for Innovation-John R. Evans Leaders Fund (CFI-JELF) program grant number 42816. Mitacs Accelerate program grant number AWD024298-IT33280. We also acknowledge the support of the Natural Sciences and Engineering Research Council of Canada (NSERC), [RGPIN-2023-03575]. Cette recherche a été financée par le Conseil de recherches en sciences naturelles et en génie du Canada (CRSNG),[RGPIN-2023-03575].


\maketitlesupplementary
\noindent
This supplementary material presents further details, including the activation function visualization, the specific hyperparameter values, the reasoning behind the proposed method, an analysis of the computational complexity of SL$^2$A, and additional applications of our approach to single image super-resolution. 




\section{Activation Function Visualization}
\begin{figure}[ht]
	\centering
	\includegraphics[width=1\linewidth]{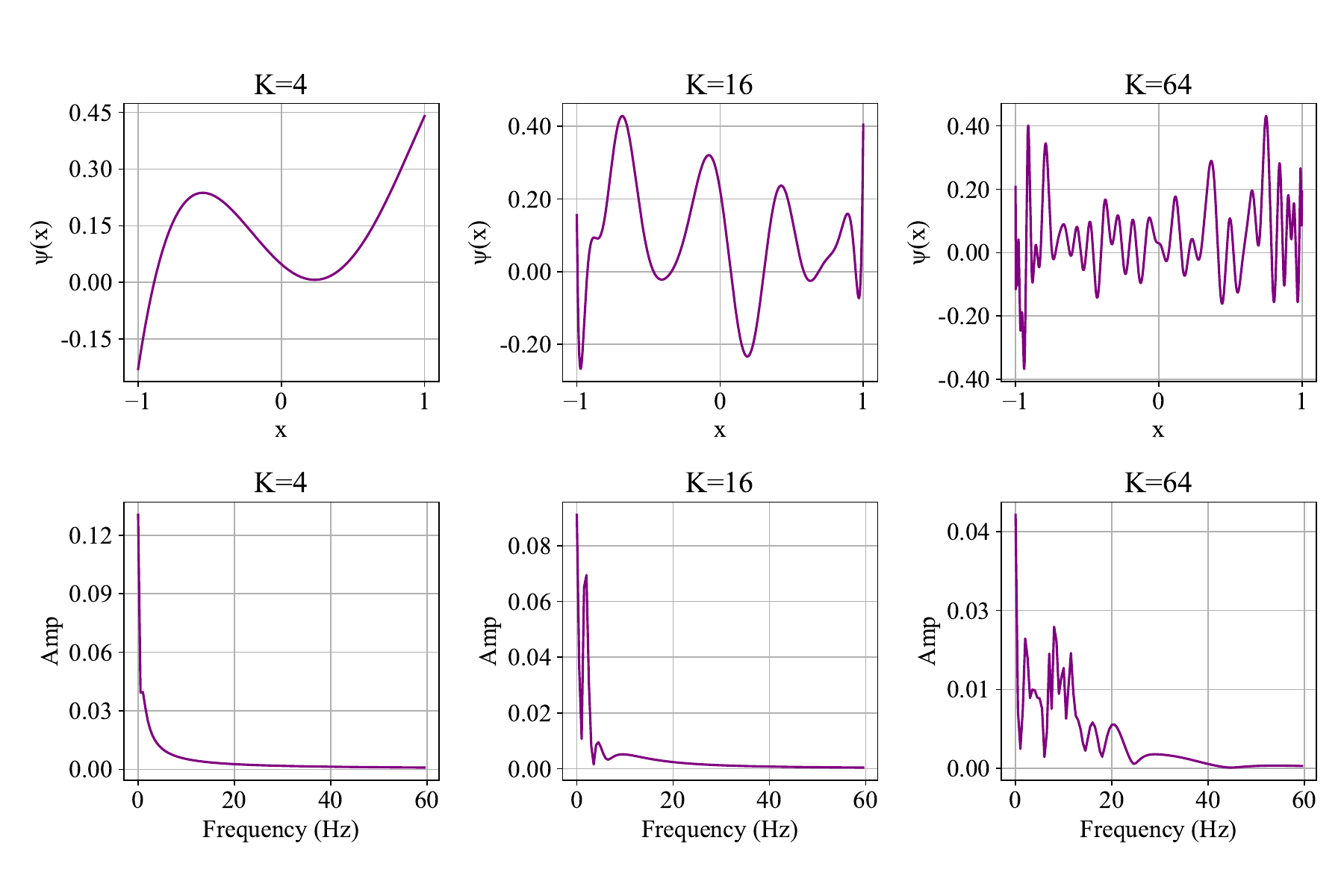}
    \caption{Frequency spectrum of our learnable activation block \( \Psi(\mathbf{x}) \).}
	\label{fig:chebyshev_plot}
\end{figure}
\noindent
\Cref{fig:chebyshev_plot} shows the frequency spectrum of learnable activation block (\( \Psi(\mathbf{x}) \)) across different values of \( K \). The top row illustrates the function \(\psi(x)\) in the time domain, while the bottom row shows its amplitude spectrum in the frequency domain. As \( K \) increases, the basis functions are capable of covering a wider frequency and bandwidth range and preserving high-frequency components, which is crucial for accurate INRs. By expanding the frequency coverage, SL$^2$A enhances the capacity to learn complex, high-frequency patterns effectively, addressing spectral bias problem.

\section{SL$^2$A Design Choice}

\subsection{Chebyshev Polynomial.}
Our architecture leverages Chebyshev polynomials to parameterize learnable functions due to their well-established advantages in numerical approximation. These polynomials exhibit superior convergence properties, numerical stability, and orthogonality, making them highly effective for function approximation \cite{rivlin1974chebyshev}.
One key motivation for using Chebyshev polynomials over B-splines is their ability to efficiently approximate activation functions such as ReLU. The minimax property of Chebyshev polynomials ensures that they minimize the maximum error in polynomial approximations, leading to higher accuracy with fewer parameters \cite{rivlin1974chebyshev,mason2002chebyshev}. 
Additionally, Chebyshev polynomials provide strong spectral approximation capabilities, making them well-suited for capturing high-frequency components of functions \cite{boyd2001chebyshev}. Compared to the B-splines used in KANs, which rely on grid-based representations and suffer from inefficiencies in deeper layers, Chebyshev polynomials offer a more flexible and efficient alternative. Prior research \cite{ss2024chebyshev,shukla2024comprehensive}, has demonstrated that integrating Chebyshev polynomials into KANs enhances efficiency, and our experiments further confirm that using them in KAN architecture leads to improved performance, as evidenced in \cref{tab:image_comparison_kan}.
While integrating Chebyshev polynomials into KANs improves efficiency and performance, the result is still suboptimal. Our approach mitigates the inefficiencies of using KANs in all layers by leveraging Chebyshev-based activation learning in earlier layers while propagating rich representations through skip connections. This hybrid approach enhances overall performance compared to standard KAN architectures. Thus, the decision to use Chebyshev polynomials instead of B-splines (or other polynomials) is driven by their superior approximation properties, stability, and efficiency, ultimately leading to improved performance in our proposed architecture. 
\begin{table}[b]
\centering
\scriptsize
\setlength{\tabcolsep}{2pt}
\renewcommand{\arraystretch}{0.8}
\vspace{-13pt}
\begin{minipage}{0.47\textwidth}
    \centering
    \caption{Comparison of PSNR of SL\textsuperscript{2}A and KAN methods for image approximation (image 00).}
    \vspace{-8pt}
    \label{tab:image_comparison_kan}
    \begin{tabular}{l |c |c| c| c c c c c}
    \toprule
      Method & \#Params (M) & Time (min.) & Size (MB) & PSNR & SSIM \\
    \midrule
       KAN (B-Spline) & 0.329 & 210.1 & 0.93 & 25.40 & 0.722\\
       KAN (Chebyshev) & \textbf{0.203} & 4.27 & \textbf{0.78} & 30.50 & 0.845\\
        SL\textsuperscript{2}A  & 0.330 & \textbf{0.77} & 0.93 & \textbf{33.40} & \textbf{0.892} \\
    \bottomrule
    \end{tabular}
\end{minipage}
\end{table}%
\noindent
\subsection{ReLU Layer.}
\begin{figure}[ht]
	\centering
	\includegraphics[width=1\linewidth]{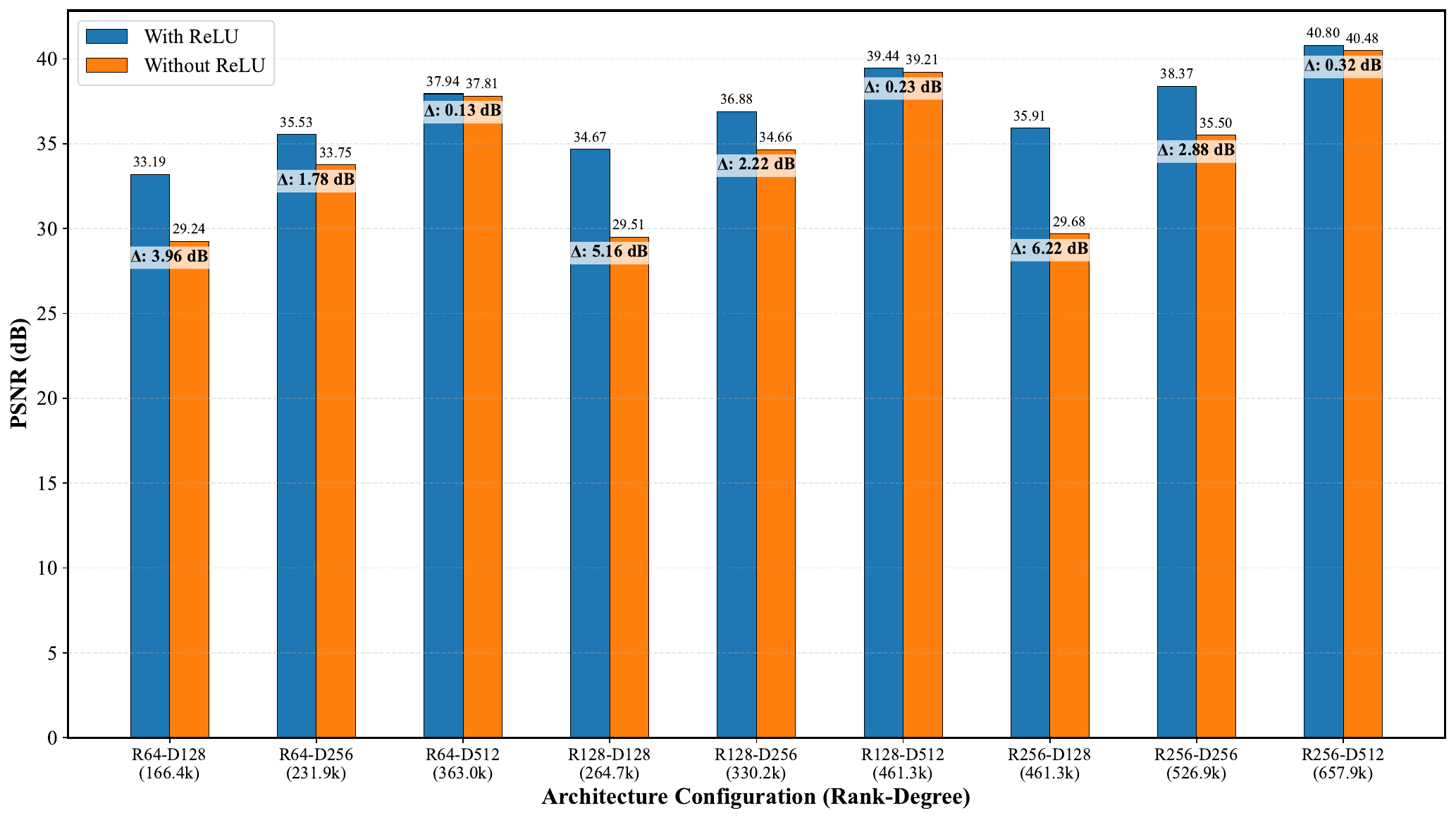}
    \caption{Comparison of PSNR (dB) across different rank-degree configurations with and without ReLU activation. Removing ReLU results in a consistent performance drop, particularly in lower-degree Chebyshev expansions, highlighting its importance in maintaining stability and expressivity.}
	\label{fig:relu-plot}
\end{figure}

\noindent
We analyze the impact of the ReLU activation in our architecture by comparing models with and without ReLU across different ranks of linear layers and polynomial degree configurations. Specifically, alternative formulations of Equation (4) in the paper could remove ReLU, leading to:


\begin{equation}
\begin{aligned}
\mathbf{z}_1 &= \Psi(\mathbf{x}), \\
\mathbf{z}_l &=  \mathbf{W}_l (\mathbf{z}_{l-1} \odot \mathbf{z}_1) + \mathbf{b}_l, \quad l = 2, 3, \ldots, L-1, \\
f_{\theta}(\mathbf{x}) &= \mathbf{W}_L (\mathbf{z}_{L-1} \odot \mathbf{z}_1) + \mathbf{b}_L.
\end{aligned}
\end{equation}
As shown in \cref{fig:relu-plot}, removing ReLU results in a consistent PSNR drop, with more severe degradation observed in lower-degree Chebyshev expansions, highlighting the critical role of ReLU in enhancing the model's expressive power to capture complex signal representations. The largest performance drop is 6.22 dB in lower-degree configurations, demonstrating its critical importance, Even in higher-degree settings, ReLU continues to provide improvements, reinforcing its necessity for expressivity. These results confirm that the modulation of layers by learnable activation \(\Psi(\mathbf{x})\) could be insufficient and introducing nonlinearity such as ReLU improves performance and expressive power.

\section{Chebyshev Degree/ Linear layer rank Efeect}

\begin{figure}[ht]
	\centering
	\includegraphics[width=1\linewidth]{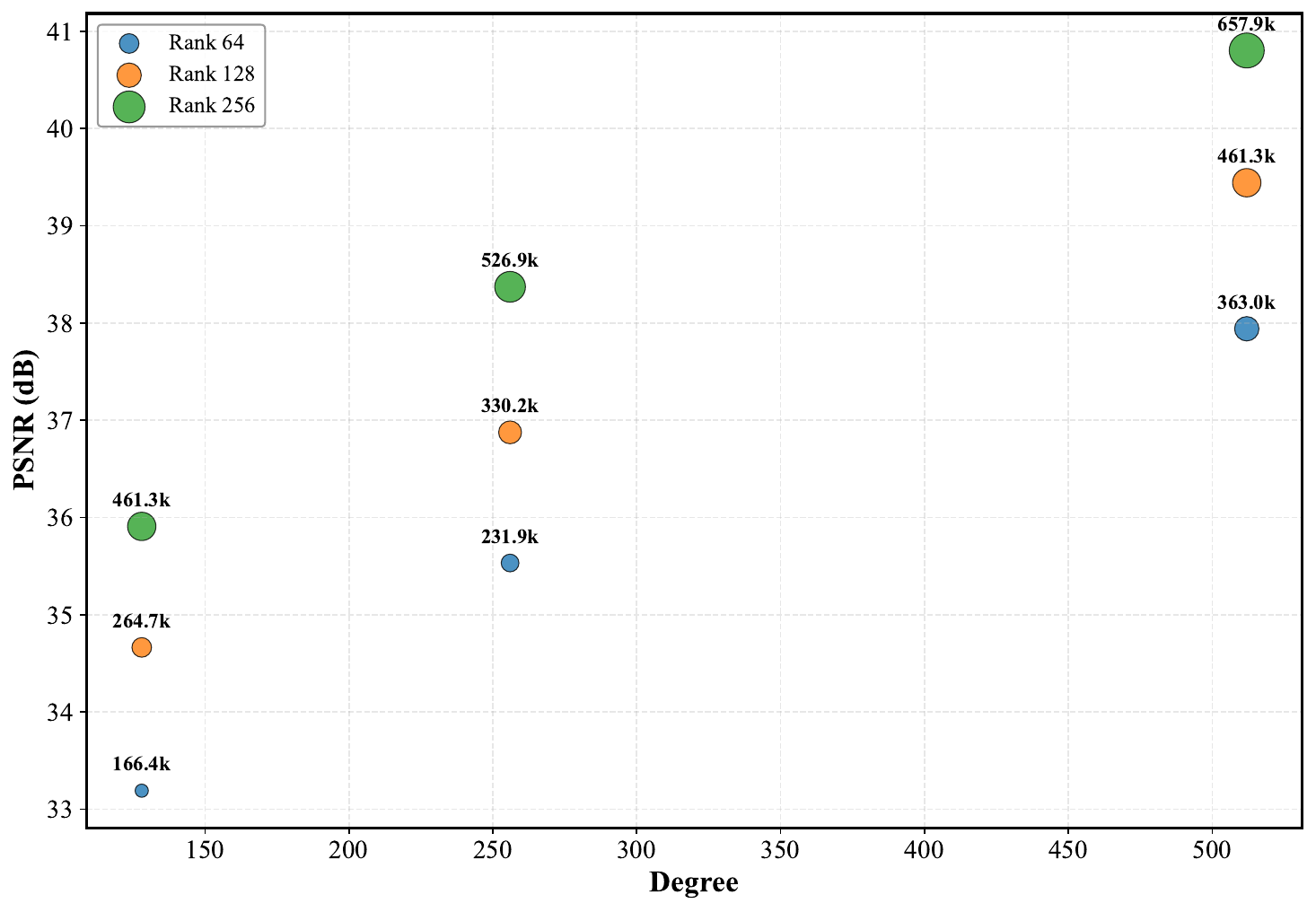}
    \caption{PSNR comparison across different rank-degree configurations and model sizes. Higher polynomial degrees lead to better performance, particularly when using lower-rank linear layers in deeper stages, supporting our architectural design choice.}
	\label{fig:degree-relu}
\end{figure}

\begin{figure}[ht]
	\centering
	\includegraphics[width=1\linewidth]{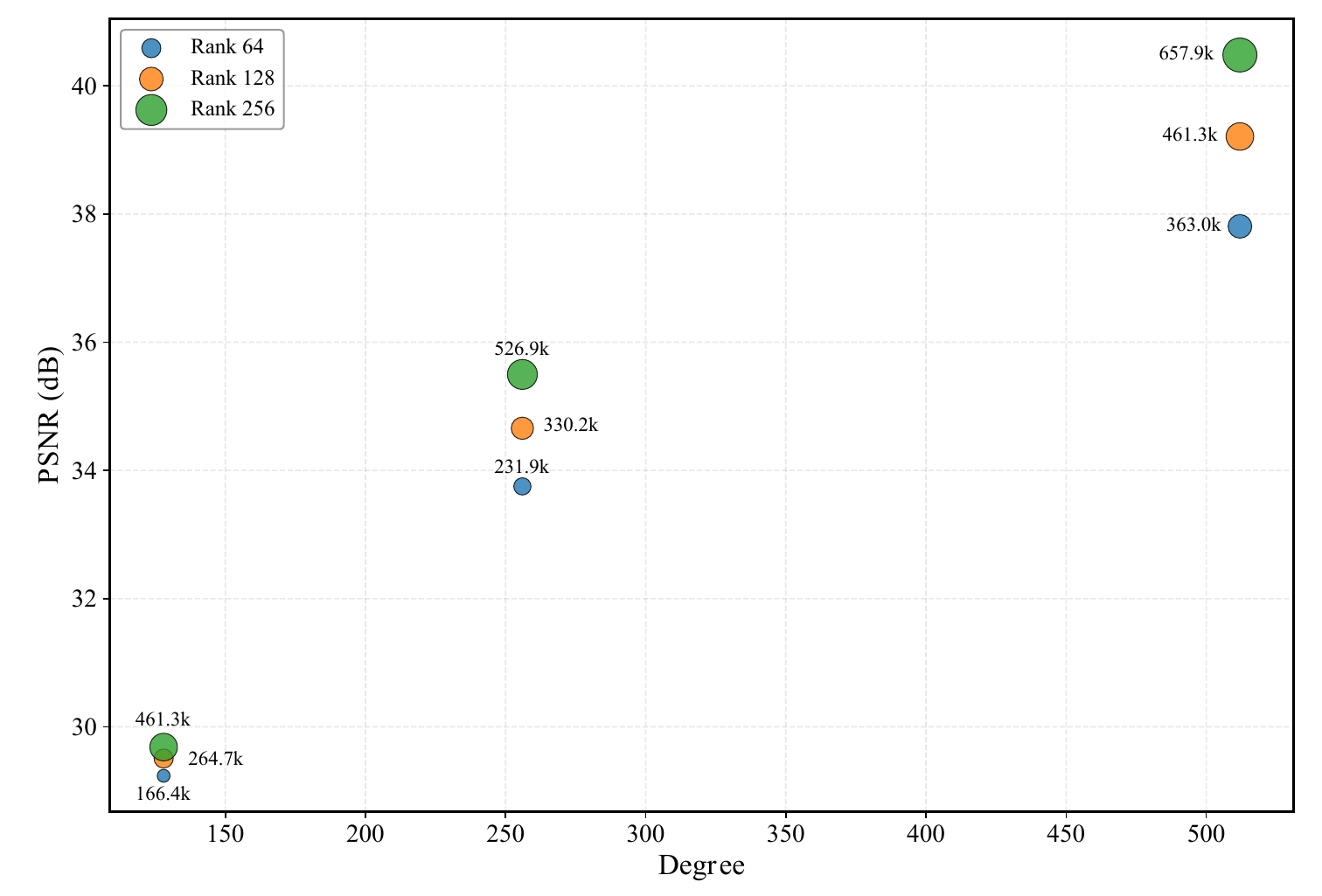}
    \caption{PSNR comparison without ReLU activation. The trend remains consistent, showing that modifying rank alone has limited effect on performance, further highlighting the importance of structured polynomial expansions.}
	\label{fig:degree-no-relu}
\end{figure}

\Cref{fig:degree-relu,fig:degree-no-relu} illustrate the impact of rank of linear layers and the polynomial degree on PSNR for image representation tasks. As shown \cref{fig:degree-relu}, increasing both the Chebyshev polynomial degree and rank of linear layers leads to higher PSNR values, demonstrating the effectiveness of higher-degree representations in capturing finer details. 
More importantly, the results reveal a critical trade-off: under similar model size or parameter constraints, increasing the polynomial degree in the initial layer generally yields better accuracy than using higher-rank configurations with a lower degree in the image representation task. For instance, the blue point on the right (363k parameters) achieves superior PSNR than the orange point in the middle (330k parameters) despite having fewer rank, and similar trends hold across other configurations. This validates our architectural choice of employing lower-rank linear layers in deeper layers while allowing a higher polynomial degree in the initial layer, enabling efficient representations without excessive parameter growth.
\Cref{fig:degree-no-relu} further supports this intuition by presenting the same analysis without ReLU activations. The observed trends remain consistent, reinforcing that altering the rank of linear layers alone is less effective in improving performance when ReLU is removed. This highlights the complementary role of non-linear activation functions and low-rank layers in achieving a balance between efficiency and expressivity in our proposed architecture.

\section{Hyperparameter Settings}
In accordance with the configurations used in prior studies, we employed the hyperparameter settings summarized in~\cref{Table:model_config} for the image fitting task. Furthermore, we adopted the same initialization scheme as in these works, utilizing the publicly available implementation provided by the authors.


\begin{table}[h!]
\centering
\caption{Configuration details of different models used in the experiments.}
\begin{tabular}{l c c c c c}
\hline
Model & \(K\) & \(\omega_0\) & \(s_0\) & Layers & Hidden Features \\
\hline
Finer & - & 30 & - & 5 & 256 \\
Wire & - & 20 & 30 & 4 & 300 \\
Gauss & - & - & 30 & 5 & 256 \\
Siren & - & 30 & - & 5 & 256 \\
ReLU+P.E. & - & - & - & 5 & 256 \\
SL$^{2}$A & 256 & - & - & 5 & 256 \\

\hline
\end{tabular}
\label{Table:model_config}
\end{table}

\section{Computational Complexity}

We analyze the computational complexity in terms of the number of parameters, and training time (per image/SDF), across different tasks, as summarized in \cref{tab:image_param_comparison,tab:occupancy_param_comparison}. Although our method, SL$^2$A, utilizes more parameters compared to FINER (the best-performing method after ours), it does not require extensive training time. It is important to note that the reported training times correspond to a fixed number of iterations using optimal hyperparameters that maximize performance (PSNR) rather than configurations optimized for speed. Time differences between methods arise from their respective optimal hyperparameter settings. Our primary contribution is the elimination of the need for manual activation function design, in contrast to prior approaches. Moreover, as illustrated in \cref{tab:image_comparison_kan}, existing methods utilizing naive learnable activations, such as KAN~\cite{liu2024kan}, suffer from limited scalability and demand substantial training resources. We effectively address these limitations by introducing a scalable learnable activation paradigm that requires only marginally increased parameter counts while maintaining computational efficiency through the use of a single learnable activation layer and low-rank MLPs, which minimize overhead without sacrificing performance. Future research will explore further enhancements aimed at improving network compactness. In the image representation task (\cref{tab:image_param_comparison}), SL$^2$A achieves competitive training time, considerably outperforming many existing methods, and remains efficient despite its marginally higher parameter count. For the occupancy representation task (\cref{tab:occupancy_param_comparison}), SL$^2$A requires slightly longer training times compared to FINER but still maintains efficiency within acceptable bounds. Overall, the complexity analysis indicates that SL$^2$A remains computationally efficient across tasks, with slight trade-offs in training time balanced by its superior performance in representation quality. 


\begin{table}[b]
\centering
\scriptsize
\setlength{\tabcolsep}{2pt}
\renewcommand{\arraystretch}{}
\begin{minipage}{0.47\textwidth}
    \centering
    \caption{Comparison of computational complexity across different methods for the Image Representation Task.}
    \vspace{-8pt}
    \label{tab:image_param_comparison}
    \begin{tabular}{l c c}
    \toprule
      Method & \#Params (M)$\downarrow$ & Training Time (min.) $\downarrow$ \\
    \midrule
       FINER & 0.198 & \textbf{0.595} \\
       WIRE & \textbf{0.099} & 1.713 \\
       Gauss & 0.198 & 3.08 \\
       SIREN & 0.198 & 0.643 \\
       ReLU+P.E. & 0.204 & 3.425 \\
       SL\textsuperscript{2}A  & 0.330 & 0.77 \\
    \bottomrule
    \end{tabular}
\end{minipage}
\end{table}


\begin{table}[b]
\centering
\scriptsize
\setlength{\tabcolsep}{2pt}
\caption{Computational complexity across different methods for the Occupancy Representation Task.}
\label{tab:occupancy_param_comparison}
\begin{tabular}{lcc}
\toprule
Method & \#Params (M)$\downarrow$ & Training Time (min.) $\downarrow$ \\
\midrule
FINER & 0.198 & 42.41 \\
WIRE & \textbf{0.066} & 63.03 \\
Gauss & 0.198 & 32.05 \\
SIREN & 0.198 & \textbf{27.22} \\
ReLU+P.E. & 0.214 & 39.32 \\
SL\textsuperscript{2}A & 0.248 & 45.59 \\
\bottomrule
\end{tabular}
\end{table}



\section{Single Image Super Resolution}
\begin{figure*}[!ht]
    \centering
    \includegraphics[width=0.8\textwidth]{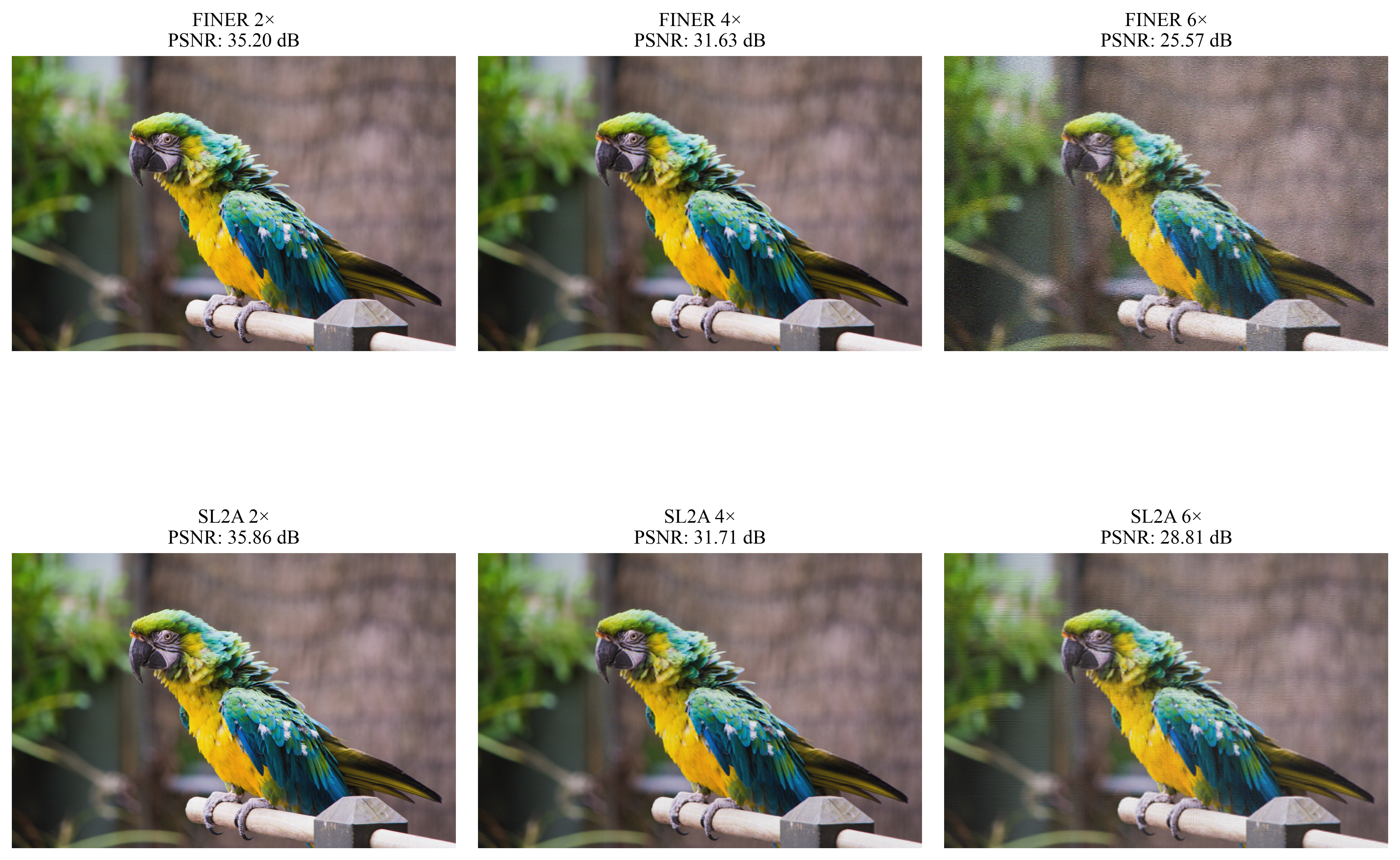}
    \caption{Results for a single image super resolution compared with FINER.}
    \label{fig:sr}
\end{figure*}

To demonstrate the generalization capability of our approach, we evaluated it on the task of single image super-resolution using an image from the DIV2K dataset~\cite{timofte2017ntire}. Specifically, we trained our architecture for image representation and downsampled the original image (with dimensions $1356 \times 2040 \times 3$) by factors of $2$, $4$, and $6$ and evaluated it on the original resolution. The corresponding results, illustrated in Fig.~\ref{fig:sr}, compare super-resolution reconstructions of a parrot image across multiple ratios. Our findings indicate that SL$^{2}$A-INR consistently surpasses FINER, achieving superior PSNR~\cite{hore2010image} metrics. Furthermore, SL$^{2}$A-INR distinctly preserves sharp details and produces less noisy result images, specifically in 6× settings, whereas the recent SOTA method FINER produces comparatively noisier results. These visual and quantitative improvements highlight our method's effectiveness in achieving high-quality super-resolution reconstructions and underscore its robustness and adaptability to inverse problems. We hypothesize that this trend could similarly hold for other inverse problem tasks involving complex reconstructions, a direction we plan to explore in future work.

\section{Initialization Scheme}

As previously mentioned, we employed the Xavier uniform initialization scheme for our method. Nevertheless, our approach (SL\textsuperscript{2}A) demonstrates robustness to various initialization schemes. To substantiate this claim, we evaluated our image representation (Image 00) using multiple initialization methods, as summarized in \cref{tab:init_comparison}. The results indicate minimal variations in performance across different initialization schemes, confirming the robustness of our method. This characteristic contrasts markedly with typical INR methods, such as SIREN \cite{sitzmann2020implicit}, whose performance significantly depends on the chosen initialization scheme, often experiencing notable degradation under suboptimal initialization.

\begin{table}[h]
\centering
\caption{Comparison of various initialization schemes evaluated on Image 00. Results demonstrate robustness of SL\textsuperscript{2}A, showing minimal variation across methods, except for the uniform initialization.}
\label{tab:init_comparison}
\begin{tabular}{lc}
\toprule
Initialization Scheme & PSNR \\
\midrule
Xavier uniform        & 33.40 \\
Kaiming uniform       & 33.22 \\
Kaiming normal        & 33.12 \\
Orthogonal            & 33.26 \\
Uniform               & 31.86 \\
Normal                & 33.28 \\
\bottomrule
\end{tabular}
\end{table}

\clearpage

{
    \small
    \bibliographystyle{ieeenat_fullname}
    \bibliography{main}
}

\end{document}